\documentclass[lettersize,journal]{IEEEtran}
\usepackage{amsmath,amsfonts}
\usepackage{algorithmic}
\usepackage{algorithm}
\usepackage{array}
\usepackage[caption=false,font=normalsize,labelfont=sf,textfont=sf]{subfig}
\usepackage{textcomp}
\usepackage{stfloats}
\usepackage{url}
\usepackage{verbatim}
\usepackage{graphicx}
\usepackage{cite}
\usepackage{booktabs}
\usepackage[normalem]{ulem}
\usepackage{xcolor}
\usepackage{multirow}
\useunder{\uline}{\ul}{}
\begin{document}

\title{AlignFreeNet: Is Cross-Modal Pre-Alignment Necessary? An End-to-End Alignment-Free Lightweight Network for Visible-Infrared Object Detection}



\author{Dingkun Zhu, Haote Zhang, Lipeng Gu, Wuzhou Quan, Fu Lee Wang, \textit{Senior Member}, \textit{IEEE},  Honghui Fan, Jiali Tang,  Haoran Xie, \textit{Senior Member}, \textit{IEEE}, Xiaoping Zhang, \textit{Fellow}, \textit{IEEE}, and Mingqiang Wei, \textit{Senior Member}, \textit{IEEE} \\

\thanks{H. Zhang, H. Fan, J. Tang, D. Zhu are with the School of Computer Science, Jiangsu University of Technology, Changzhou, China (e-mail: zhangrichard168@gmail.com; fanhonghui@jsut.edu.cn; tangjl@jsut.edu.cn; zhudingkun@jsut.edu.cn).}
\thanks{L. Gu, W. Quan and M. Wei are with the School of Computer Science and Technology, Nanjing University of Aeronautics and Astronautics, Nanjing, China (e-mail: glp1224@163.com; q.wuzhou@gmail.com; mingqiang.wei@gmail.com).}
\thanks{F. Wang is with the School of Science and Technology, Hong Kong Metropolitan University, Hong Kong SAR (e-mail: pwang@hkmu.edu.hk).}
\thanks{H. Xie is with the School of Data Science, Lingnan University, Hong Kong, China (e-mail: hrxie@ln.edu.hk).}
\thanks{Xiao-Ping Zhang is with the Tsinghua Shenzhen International Graduate School, Tsinghua University, Shenzhen 518055, China (e-mail: xpzhang@ieee.org).}
\thanks{\textit{(Corresponding author: Mingqiang Wei.)}}
}

\maketitle

\begin{abstract}

Cross-modal misalignments, such as spatial offsets, resolution discrepancies, and semantic deficiencies, frequently occur in visible-infrared object detection (VI-OD).
To mitigate this, existing methods are typically adapted into an alignment-based fusion paradigm, in which an explicit pixel- or feature-level alignment module is inserted before cross-modal fusion.
However, pixel-level alignment struggles to cope with severe or mixed misalignments, whereas feature-level alignment often introduces undesirable noise into fused representations under such conditions, ultimately limiting detection performance.
In this paper, we propose a novel alignment-free network (AlignFreeNet) for VI-OD.
Differing from prior methods, AlignFreeNet abandons any explicit alignment and instead adopts an alignment-free fusion paradigm.
Specifically, AlignFreeNet comprises two core modules: variation-guided cross-modal compensation (VCC) and frequency-guided cross-modal fusion (FCF).
VCC adaptively feeds the compensated information derived from cross-modal discrepancies back into each modality, enhancing visible and infrared representations without the noise caused by explicit alignment.
FCF achieves robust cross-modal fusion by suppressing task-irrelevant redundancy via frequency-domain gating, effectively mitigating noise introduced in the process.
Moreover, VCC and FCF jointly exploit low- and high-frequency cues to preserve foreground contours in fused representations, effectively mitigating cross-modal blending caused by severe mixed misalignments.
Extensive evaluations on DVTOD, M$^3$FD, and DroneVehicle demonstrate that our AlignFreeNet achieves state-of-the-art performance under severe mixed misalignment conditions, highlighting its robustness and generalization.


\end{abstract}


\section{Introduction}
\IEEEPARstart{V}{isible}-infrared object detection (VI-OD) leverages complementary information from visible and infrared modalities to achieve robust environmental perception, and plays a key role in applications such as remote sensing, autonomous driving, and embodied intelligence \cite{zou2023object,liu2024infrared}.
However, frequent misalignments between the visible and infrared modalities continue to pose a significant challenge for VI-OD.

\begin{figure}[!t]
    \centering
    \vspace{0cm}
    \includegraphics[trim={255 80 262 90},clip,width=\linewidth]{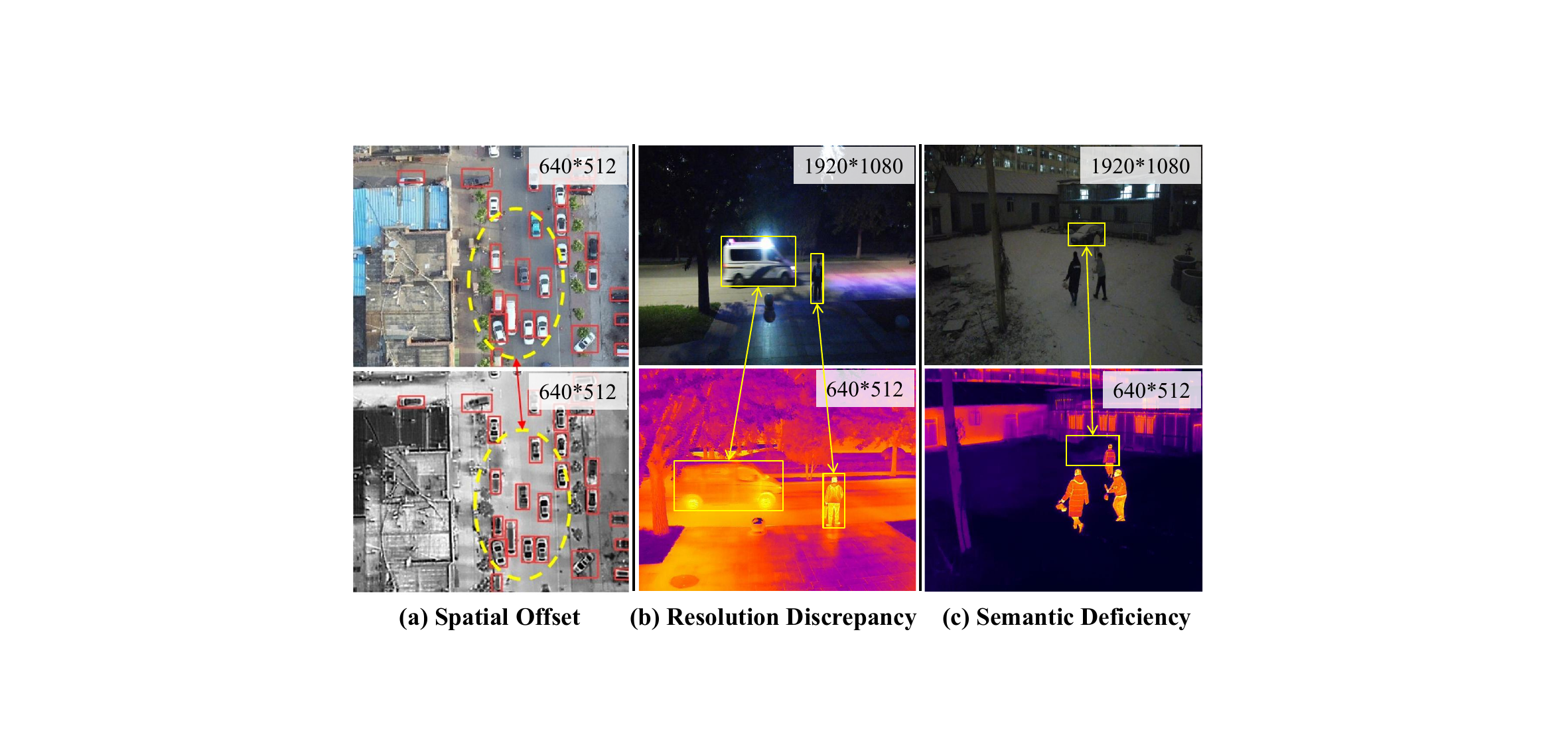}
    \vspace{-8 mm}
    \caption{
    \textbf{Illustration of cross-modal misalignments.}
    Common misalignments include: (a) \textit{spatial offsets}, caused by variations in acquisition conditions such as capture angle and timing; (b) \textit{resolution discrepancies}, arising from differences in sensor resolutions and focal lengths; and (c) \textit{Semantic Deficiencies}, stemming from inherent inconsistencies between visible and infrared spectra.
    }
    \label{intro}
\end{figure}

Common cross-modal misalignments can be categorized into three major types: spatial offsets, resolution discrepancies, and semantic deficiencies.
As illustrated in Fig. \ref{intro}, spatial offsets arise from variations in acquisition conditions, such as sensor viewpoints, capture angles, and timing, which lead to misaligned object locations across modalities. 
Resolution discrepancies result from differences in sensor resolutions, focal lengths, or optical properties, causing inconsistencies in object sizes, shapes, and feature granularity. 
Semantic deficiencies stem from inherent differences between visible and infrared spectra, where certain objects may be observable in only one modality or exhibit significantly different textures.

\begin{figure*}[t!]
    \centering
    \includegraphics[trim={430 450 500 520},clip,width=\textwidth]{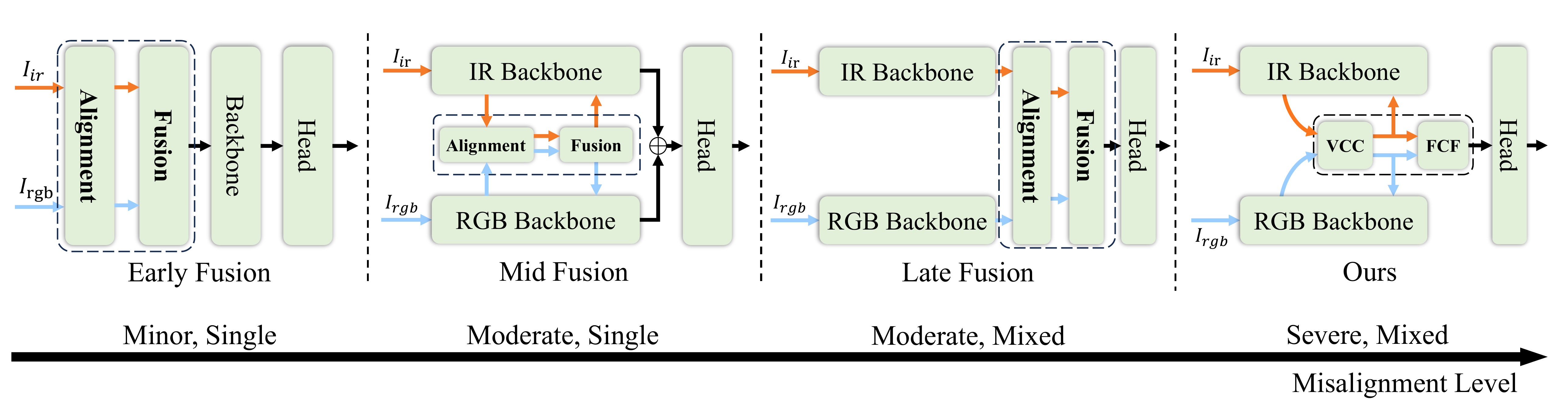}
    \caption{
    \textbf{Illustration of our method compared with the three existing fusion paradigms.} 
    Existing fusion methods typically adopt an alignment-based paradigm, in which pixels or features are explicitly aligned before cross-modal fusion.
    In contrast, our method leverages modality-variation- and wavelet-guided cues to achieve an \textbf{alignment-free fusion} in the mid-to-late fusion stages, effectively addressing mixed types of misalignment.
    This design eliminates the dependence on explicit alignment modules while substantially enhancing robustness.
    }
    \label{modes}
\end{figure*}

To mitigate the above cross-modal misalignments, existing VI-OD methods adopt different strategies depending on their fusion paradigms (i.e., early, mid, or late fusion).
As illustrated in Fig. \ref{modes}, early-fusion methods~\cite{liu2022target,zhang2023superyolo,xu2023murf,zhao2024equivariant,medeiros2024mipa,li2025mulfs,li2025fd2} align modalities at the pixel level by leveraging geometric transformations, feature matching, or resampling strategies. 
They can effectively handle minor single misalignments, such as small viewpoint changes and limited geometric distortions. However, they struggle to cope with severe misalignments involving significant cross-modal discrepancies, local deformations, or mixed misalignments.
Mid-fusion methods~\cite{qingyun2021cross,yuan2024c,yuan2024improving,liu2024cross,dong2024fusion,zhao2025reflectance,zhu2025wavemamba,liu2025retinexdet} align shallow or intermediate features through deformable convolutions, attention-based warping, or cross correlation. 
Although these methods can effectively handle moderate single misalignments and enhance the representation of small objects, their cross-modal correspondences become ambiguous and unreliable under mixed misalignments. Consequently, alignment at this stage amplifies uncertainty, introduces misleading features and structural noise, and incurs considerable computational overhead.
Late-fusion methods~\cite{zhang2019weakly,cao2023multimodal,shen2024icafusion,zhou2025dmm,qu2025gcmf,zheng2025gaanet} align deep semantic features using cross-attention mechanisms and shared embedding spaces. 
Although these methods leverage deep semantic consistency to compensate for spatial and resolution misalignments, they are particularly susceptible to semantic deficiencies (see Fig. \ref{intro} (c)), which may result in weak or even false cross-modal correspondences. Such effects are especially detrimental to small objects, ultimately hindering their accurate recognition.
Overall, existing methods follow an alignment-based fusion paradigm, assuming that aligning visible and infrared modalities ensures effective feature fusion. Yet, when facing severe or mixed misalignments, this assumption fails, introducing uncertainty and noise that ultimately degrade detection performance.

The above limitations naturally lead to the question of \textit{whether modality alignment before fusion is indeed indispensable for addressing scenarios with cross-modal misalignments}.
To this end, we introduce AlignFreeNet, a novel alignment-free network for VI-OD, which is innovatively built upon the alignment-free fusion paradigm. AlignFreeNet combines the advantages of mid-level and late fusion: it captures cross-modal discrepancies at the mid-fusion stage to enhance modality complementarity, and further performs efficient late fusion to suppress noise interference on foreground contours, thereby addressing severe mixed misalignments within a unified framework.
Specifically, it comprises two key modules:
i) Variation-guided cross-modal compensation (VCC). Instead of directly learning inter-modal correlations for cross-modal alignment, VCC emphasizes modality-specific variations during mid-level fusion and adaptively reintegrates them into each modality, thereby enhancing both visible and infrared representations and avoiding noise interference caused by misalignment.
ii) Frequency-guided cross-modal fusion (FCF). 
FCF employs region- and channel-wise frequency gating to suppress redundant cross-modal information, effectively mitigating noise introduced in the process. Moreover, FCF leverages both low- and high-frequency cues to enhance the integrity of foreground contours, effectively mitigating cross-modal blending caused by severe mixed misalignments.

In summary, our main contributions are:
\begin{itemize}
    \item We propose AlignFreeNet, a novel alignment-free network for VI-OD that addresses mixed misalignments, including spatial offsets, resolution discrepancies, and semantic deficiencies. It features VCC and FCF modules.
    \item VCC captures modality-specific variations during mid-level fusion and adaptively propagates them across modalities, effectively mitigating noise from direct feature-level alignment.

    \item FCF mitigates misalignment-induced fusion errors by employing region–channel-wise frequency gating to suppress redundant cross-modal information, effectively leveraging frequency cues to preserve clear target boundaries under misaligned overlaps.
\end{itemize}

\section{Related Work}
\subsection{Cross-Modality Object Detection}
Single-modality data often appears to be highly susceptible to environmental distractions, etc. Cross-modality object detection can compensate by leveraging complementary information from heterogeneous sensors (e.g., RGB cameras, LiDAR, and Infrared imaging) to improve detection robustness under challenging conditions. Fusion as the main technique for exploiting cross-modality pairs~\cite{bayoudh2022survey} was focused on by many studies to improve detection, Transformer, with its powerful modeling capacity, has been used in a large number of studies of fusion methods~\cite{qingyun2021cross,yuan2024improving,shen2024icafusion}, Fang et al.~\cite{qingyun2021cross} first proposed a Transformer-based fusion method for object detection, which made significant progress in the acquisition of complementary information. 
An efficient new module proposed by Gu et al. [mamba~\cite{gu2023mamba,dao2024transformers}] was also adopted by several methods~\cite{li2024cfmw,shen2025htd}. Despite the importance of fusion methods, the task still faces other challenges that need to be addressed, such as image degradation caused by low light, haze, noise, snow, and rain, as well as misalignment of image pairs. Recently, a large number of studies have emerged to address these challenges associated with the task. Zhang et al.~\cite{zhang2023illumination} proposed a light-guidance method for fusion of inter- and intra-modal information in low-light detection. Li et al.~\cite{li2024cfmw} using a diffusion model and mamba-based fusion to solve the adverse weather problem. 

Although significant progress has been made in advancing fusion techniques and mitigating image degradation, the issue of multispectral misalignment remains unavoidable. Existing methods often rely on manually aligned data and fail to account for real-world scenarios. In contrast, our proposed approach adaptively addresses various types of misalignment.

\begin{figure*}[htb]
    \centering
    \includegraphics[trim={760 200 760 200},clip,width=\textwidth]{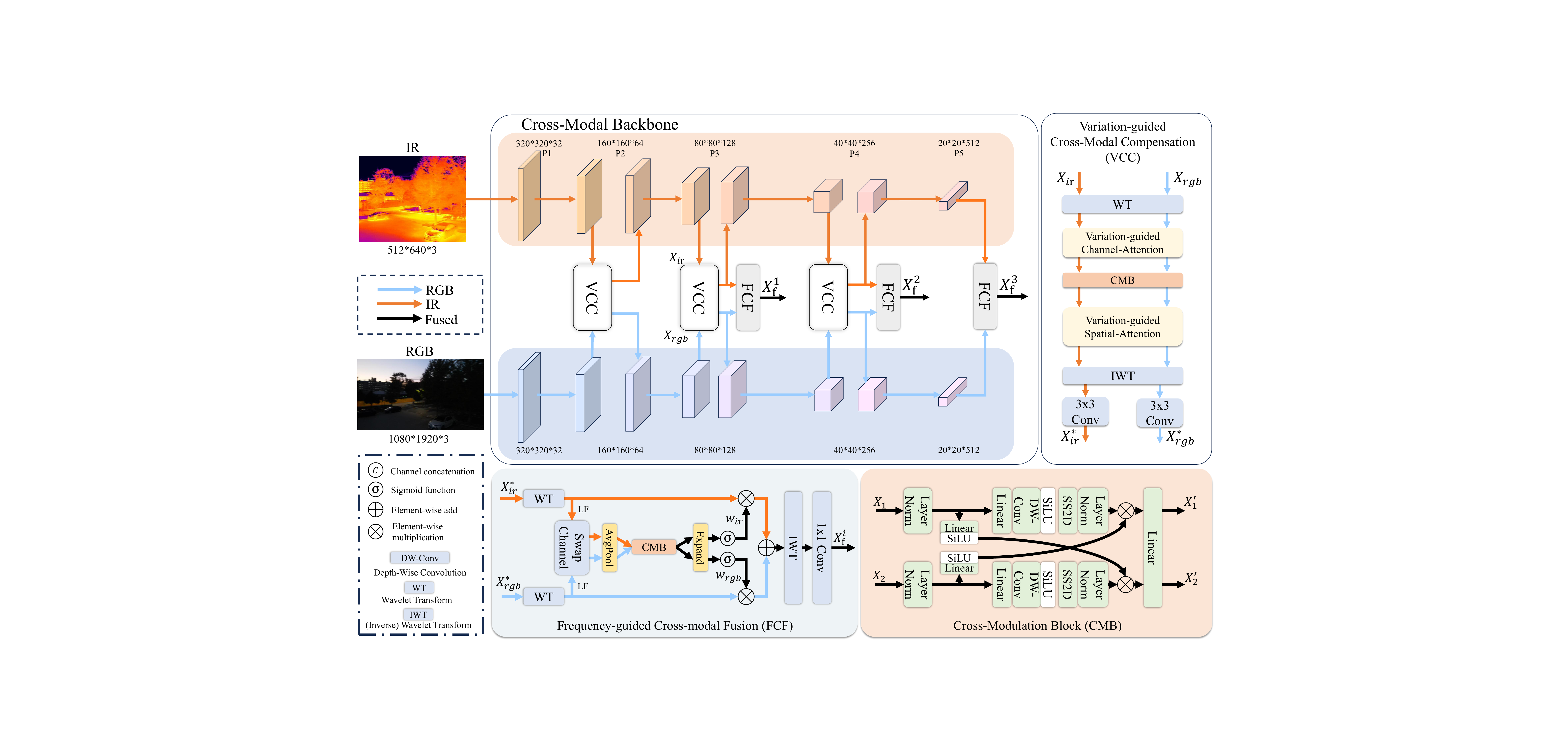}
   \vspace{-8 mm}
    \caption{The overall structure of our proposed method, alignment-free network (AlignFreeNet). There are three separate VCC and FCF in the backbone for feature fusion. In VCC, we conduct the major cross-modality interaction of our network based on modality variation, and the specific algorithm is provided in \textbf{Section C}. In FCF, we utilize the variation-enhanced feature from VCC as the basis for selectively fusing features through a frequency-domain gated mechanism. Outputs from VCC are sent back to the backbone, and the outputs from FCF are used as the inputs to the detection head. \label{fig-flow}}
    \label{method}
\end{figure*}

\subsection{Cross-Modality Alignment}
Image alignment transforms a source image to match a reference image, aligning them to the same coordinate system for pixel-level correspondence. However, pixel-level image alignment methods~\cite{wang2021robust,xu2023murf,li2024deep,li2025mulfs} in the early fusion stage are vulnerable to severe or mixed misalignments, as they commonly rely on explicit alignment algorithms, and they tend to be computationally expensive, as they aim to reconstruct images and preserve non-semantic details for visual effect. VI-OD mainly focuses on feature-level alignment in the mid or late stage to effectively integrate complementary target information to enhance detection performance. 
To address spatial offset misalignment,~\cite{zhang2019weakly} introduced region feature alignment (RFA) to predict the positional difference of the same object across the two modalities. Yuan et al.~\cite{yuan2022translation} proposed a translation-scale-rotation alignment module to predict the offset of objects in terms of position, size, and angle. Liu et al.~\cite{liu2024cross} proposed an offset-guided fusion mechanism that utilizes different scale levels of features to guide the light offset problem. These methods typically rely on warping or a multi-level structure to mitigate the impact of light spatial offset.
To address resolution misalignment, Song et al.~\cite{song2024misaligned} proposed a feature-level matcher to align images in the DVTOD dataset they introduced. 
Zhao et al.~\cite{zhao2025reflectance} utilize deformable convolutions to do reflectance-guided remapping to address different misalignment problems.
Zheng et al.~\cite{zheng2025gaanet} proposed an alignment method based on graph aggregation.
These methods rely on attention correlation or deformable convolutions to mitigate the impact of spatial offset mixed with resolution discrepancy.
To solve semantic deficiency misalignment, Shen et al.~\cite{shen2024icafusion} proposed an iterative fusion strategy that iteratively enhances feature representation and avoids interfering with the backbone features to reduce the impact of misalignment. Although this method is effective for single-type misalignment, its performance remains limited when dealing with mixed-type misalignments due to its reliance on semantic tolerance.

Although current methods have made tremendous progress in misaligned VIOD, methods based on feature warping, shifting, matching, etc, are often ineffective when facing diverse misalignment scenarios. Methods that are based on correlation or feature deformation often introduce ambiguity into feature representation. In contrast, our method enables adaptation to various misalignments in an alignment-free paradigm.

\section{Method}


This section introduces the Alignment-Free Network (AlignFreeNet), which establishes an alignment-free fusion paradigm for visible–infrared object detection. Built upon a dual-stream YOLO backbone (Fig.~\ref{fig-flow}), AlignFreeNet leverages cross-modal variations and wavelet-domain guidance to address severe mixed misalignments. It comprises two core components, VCC and FCF, whose collaborative multi-level fusion structure effectively handles severe mixed misalignment.

\begin{figure}[!t]
    \centering
    \vspace{0cm}
    \includegraphics[trim={100 100 100 100},clip,width=\linewidth]{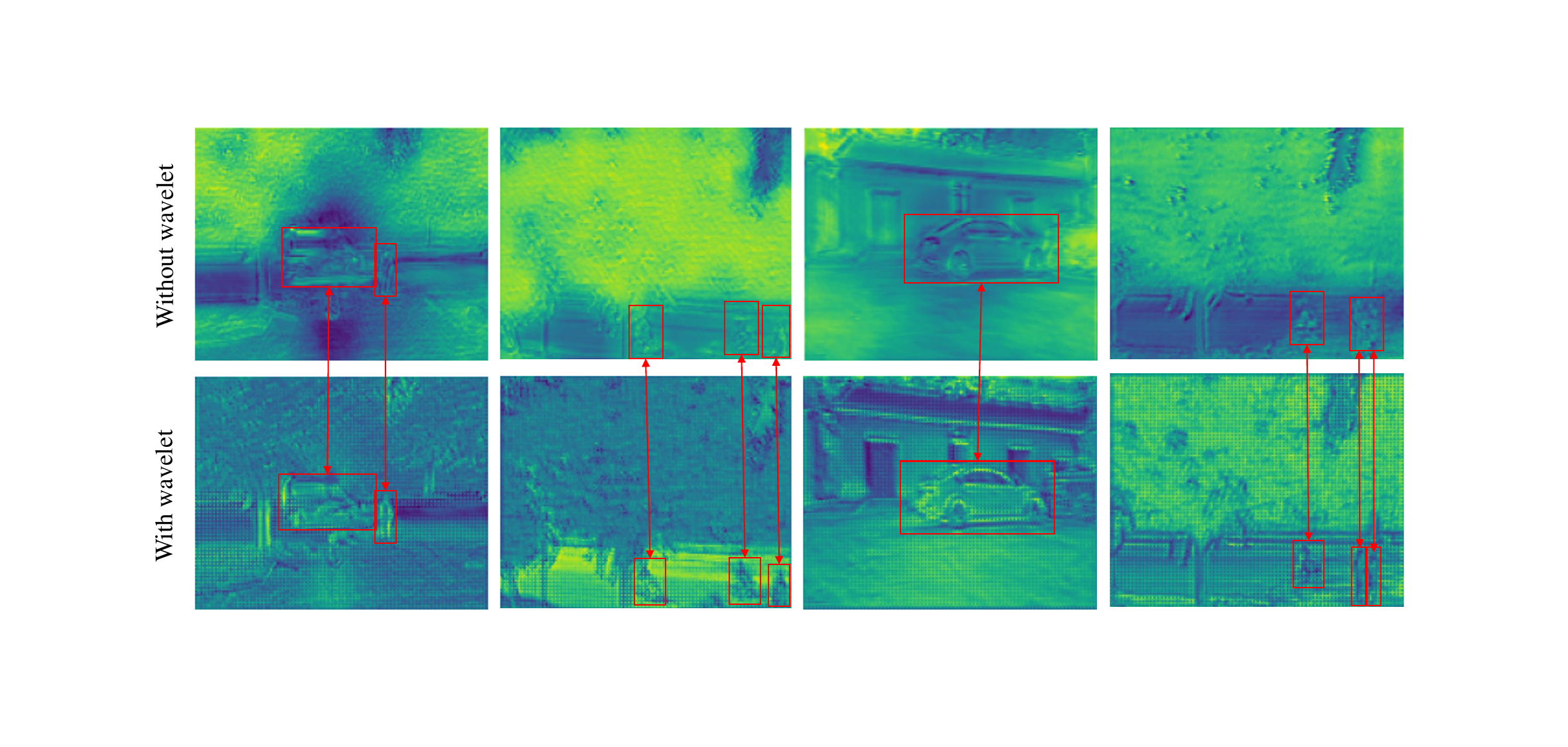}
    \caption{The difference of our model w/ and w/o wavelet transform processing. Our wavelet structure prevents targets from merging into the unrelated background or other targets caused by misalignment.}
    \label{wavefor}
\end{figure}

\subsection{Preliminary}
\textbf{Haar Wavelet Transform.} 
Given a feature map $\mathbf{X}$, a depth-wise convolution with a stride of $2$ is apply to $\mathbf{X}$ using the low-pass filter $L=\frac{1}{\sqrt{2}}[1,1]$ and the high-pass filter $H=\frac{1}{\sqrt{2}}[1,-1]$. 
This produces
\begin{equation}
\mathbf{X}_{LF},\mathbf{X}_{HF}=WT(\mathbf{X}),
\mathbf{X}=IWT(\mathbf{X}_{LF},\mathbf{X}_{HF}),
\end{equation}
where $WT(\cdot)$ denotes Haar wavelet transform.
The low-frequency subband $\mathbf{X}_{LF}=\{\mathbf{X}_{LL}\}$ is used to capture coarse information, while the high-frequency subbands $\mathbf{X}_{HF}=\{\mathbf{X}_{LH},\mathbf{X}_{HL},\mathbf{X}_{HH}\}$ are used to encode fine-grained details.
The inverse Haar wavelet transform $IWT(\cdot)$ is then applied to reconstruct the original feature map from the subbands.

\textbf{2D Selective Scan (SS2D)} \cite{liu2024vmamba}. 
It adapts state-space modeling to vision tasks by bridging the gap between 2D images and 1D sequence processing. Instead of directly applying the selective scan method as in Mamba~\cite{gu2023mamba}, image patches are expanded along four directions to form independent sequences. Each sequence is then processed using the Selective Scan State Space Model (S6) \cite{gu2023mamba}, and the resulting outputs are aggregated to reconstruct the original 2D feature map.

\subsection{Overview}
To overcome the limitations of the alignment-based fusion paradigm, we propose a lightweight, multi-level alignment-free framework for visible-infrared object detection, integrating variation-guided cross-modal compensation (VCC) and frequency-guided cross-modal fusion (FCF). VCC focuses on shallow and intermediate feature stages to capture modality-specific variations and adaptively compensate weak target representations, which is particularly effective in mitigating the impact of semantic deficiencies by enhancing the discriminative cues within each modality and guiding their complementary interaction.
In contrast, FCF operates at deeper stages, where frequency-domain gating provides robustness against spatial and resolution misalignments, selectively suppressing task-irrelevant redundancy while preserving foreground structural contours.
Multiple VCC and FCF modules are embedded into the backbone. For instance, the second VCC receives feature maps $\mathbf{X}_{ir}$ and $\mathbf{X}_{rgb}$, performs cross-modal compensation, and produces refined outputs $\mathbf{X}_{ir}^{*}$ and $\mathbf{X}_{rgb}^{*}$, which are fed back into the backbone. The corresponding FCF then fuses these enhanced features to produce $\mathbf{X}_f^{2}$. Outputs from all FCFs, $\mathbf{X}_f^{1}$, $\mathbf{X}_f^{2}$, and $\mathbf{X}_f^{3}$, are used as multi-level inputs to the YOLO detection head.
By progressively coupling VCC and FCF across multiple levels, our framework jointly addresses semantic and geometric misalignments, preserves foreground structures, reinforces weak target representations, and suppresses structural noise from misaligned cross-modal features, achieving stable and efficient cross-modal fusion across diverse misalignment scenarios.

\subsection{Variation-guided 
Cross-Modal Compensation
}
To address complex and coexisting misalignment in VI-OD, we propose the Variation-guided  Cross-Modal Compensation (\textbf{VCC}) module. Since the correspondence between misaligned modalities is often weak, directly learning inter-modal correlations is suboptimal, as ambiguity can distort feature fusion. Instead, we enable the model to learn modality-specific variations during the mid-fusion process, thereby integrating misalignment awareness directly into the fusion procedure to help improve weak target representation rather than applying a separate alignment before fusion.

In the $i^{th}$ VCC as shown in Fig.~\ref{method}, we first decompose $\mathbf{X}_{ir}$ and $\mathbf{X}_{rgb}$ using the wavelet transform. This decomposition allows VCC to separately handle coarse structures and fine-grained details, ensuring both global and local target information are enhanced, and then we obtain the modality-variation to perform the \textbf{Variation-based Channel-Attention}:
\begin{equation}
\begin{array}{c c c c}
\mathbf{X}_{LF}^{ir},\mathbf{X}_{HF}^{ir}=WT(\mathbf{X}_{ir}^{*}),\mathbf{X}_{LF}^{rgb},\mathbf{X}_{HF}^{rgb}=WT(\mathbf{X}_{rgb}^{*})\\[6pt]
\mathbf{X}_{F}^{ir}=\mathbf{X}_{LF}^{ir}+\mathbf{X}_{HF}^{ir},\mathbf{X}_{F}^{rgb}=\mathbf{X}_{LF}^{rgb}+\mathbf{X}_{HF}^{rgb}\\[6pt]
\mathbf{X}_{1}^{ir} = Concat(\mathbf{X}_{F}^{rgb}-ReLU(\mathbf{X}_{F}^{ir}),\mathbf{X}_{F}^{ir})\\[6pt]
\mathbf{X}_{2}^{rgb} = Concat(\mathbf{X}_{F}^{ir}-ReLU(\mathbf{X}_{F}^{rgb}),\mathbf{X}_{F}^{rgb})
\end{array},
\end{equation}
where $Concat(\cdot)$ denotes channel-wise concatenation and $ReLU(\cdot)$ is the application of the linear rectification function. This channel-wise attention identifies the most informative modality-specific variations to reinforce weak targets, while their integration with intra-modality channels effectively suppresses irrelevant features.
We then fed $\mathbf{X}_{1}^{ir}$ and $\mathbf{X}_{2}^{rgb}$ into our CMB module shown in Fig.~\ref{method} to learn and select the modality variation to enhance:
\begin{equation}
\begin{aligned}
&\mathbf{X}_{dif}^{ir},\mathbf{X}_{dif}^{rgb}=CMB(\mathbf{X}_{1}^{ir},\mathbf{X}_{2}^{rgb})\\
\end{aligned},
\end{equation}
where $CMB(\cdot)$ denotes the application of the CMB module, $\mathbf{X}_{dif}^{ir}$ and $\mathbf{X}_{dif}^{rgb}$ are the learned variations.
We then element-wise add the learned variation to the Low-Frequency (LF) feature map and enhance the similarity based on cosine similarity (\textbf{Variation-guided Spatial Attention}):
\begin{equation}
\begin{array}{c c c c c}
\mathbf{X}_{ir}^{'}=Concat(\mathbf{X}_{LF}^{ir}+\mathbf{X}_{dif}^{ir},\mathbf{X}_{HF}^{ir})\\[9pt]
\mathbf{X}_{rgb}^{'}=Concat(\mathbf{X}_{LF}^{rgb}+\mathbf{X}_{dif}^{rgb},\mathbf{X}_{HF}^{rgb})\\[12pt]
\mathrm{Corr}(\mathbf{X}_{ir}^{\prime}, \mathbf{X}_{rgb}^{\prime}) 
= \mathrm{ReLU}\!\left( 
\frac{\mathbf{X}_{ir}^{\prime} \cdot \mathbf{X}_{rgb}^{\prime}}
{ \lVert \mathbf{X}_{ir}^{\prime} \rVert \, \lVert \mathbf{X}_{rgb}^{\prime} \rVert }
\right) \\[12pt]
\mathbf{X}_{ir}^{\prime\prime} 
= \mathbf{X}_{ir}^{\prime} 
+ \mathbf{X}_{rgb}^{\prime} \ast 
\mathrm{Corr}(\mathbf{X}_{ir}^{\prime}, \mathbf{X}_{rgb}^{\prime}) \ast s \\[9pt]
\mathbf{X}_{rgb}^{\prime\prime} 
= \mathbf{X}_{rgb}^{\prime} 
+ \mathbf{X}_{ir}^{\prime} \ast 
\mathrm{Corr}(\mathbf{X}_{ir}^{\prime}, \mathbf{X}_{rgb}^{\prime}) \ast s
\end{array},
\end{equation}
where $Corr(\cdot)$ represents cosine similarity and $s$ is a scaling factor that is set to 0.3. This cosine-similarity reinforcement selectively amplifies learned modality-specific variations based on cross-modal consistency in the frequency domain, highlighting spatial regions that are semantically or structurally important. This mechanism enhances informative features for downstream fusion, without enforcing explicit alignment between modalities. Subband in $\mathbf{X}_{ir}^{''}$ and $\mathbf{X}_{rgb}^{''}$ are then composed:
\begin{equation}
\begin{array}{l}
\mathbf{X}_{ir}^{*} = Conv_{3\times3}(IWT(\mathbf{X}_{ir}^{''}))\\[3pt]
\mathbf{X}_{rgb}^{*} = Conv_{3\times3}(IWT(\mathbf{X}_{rgb}^{''}))
\end{array},
\end{equation}
where $Conv_{3\times3}(\cdot)$ denotes $3\times3$ convolution, the convolution is used to aggregate features while preserving local details; larger kernels could overly smooth small-target features, reducing their discriminability. $\mathbf{X}_{ir}^{*}$ and $\mathbf{X}_{rgb}^{*}$ are the outputs of the module.

\subsection{Cross-Modulation Block}
To further enhance cross-modal feature interaction while suppressing misaligned representations, we construct a variant of the VSS block \cite{liu2024vmamba}, the Cross-Modulation Block (CMB), as illustrated in Fig.~\ref{method}. Unlike conventional cross-attention mechanisms that rely on explicit correspondence, the cross-modulation strategy incorporates inter-modal information as a soft gating signal, enabling more flexible and robust interaction under severe misalignment.
Specifically, the inputs $\mathbf{X}_1$ and $\mathbf{X}_2$, are linearly projected and split into two parallel streams: one serves as the representation branch for learning, and the other acts as the gating branch to regulate the complementary modality.
The dual-branch design allows the representation branch to preserve modality-specific information, while the gating branch dynamically suppresses task-irrelevant activations, thereby enhancing the discriminative components that are consistent across modalities.
The representation branch is sequentially processed by a $3\times3$ depth-wise convolution (DW-Conv), SiLU activation, the core SS2D module, and layer normalization (LayerNorm), whereas the gating branch only passes through a SiLU activation. Finally, the two cross-modally gated representations are combined channel-wise and passed through a shared Linear layer to promote modality-wise synergy, after which they are split to produce the outputs $\mathbf{X}_1^{'}$ and $\mathbf{X}_2^{'}$. The application of the module can be formulated as:
\begin{equation}
\mathbf{X}_{1}^{'},\mathbf{X}_{2}^{'}= CMB(\mathbf{X}_{1},\mathbf{X}_{2}),
\end{equation}

\subsection{Frequency-guided Cross-modal Fusion }
Since cross-modality feature representations misalign spatially, directly fusing them often amplifies interference and ambiguity due to weak correspondences. To address this issue, we propose a Frequency-guided Cross-modal Fusion  (\textbf{FCF}) module, which selectively suppresses redundant features in the frequency domain. Unlike spatial-domain features, frequency-domain representations, especially low-frequency components, are inherently less sensitive to local shifts or minor deformations, providing a more stable basis for cross-modal fusion. In addition, regional attention provides a coarse yet structurally stable weighting scheme, which is less sensitive to local misalignments compared with fine-grained pixel-level attention. Cross-modal relationships are often more reliable at the object or region level rather than at exact pixel locations, and this coarse aggregation enables the model to emphasize semantically relevant areas while suppressing misaligned or noisy features. Combined with frequency-domain representations, which are inherently less sensitive to local spatial shifts, this design offers a stable foundation for alignment-free cross-modal fusion.

Specifically, in the $i^{th}$ layer, the input features $\mathbf{X}^*_{ir}$ and $\mathbf{X}^*_{rgb}$ are first decomposed into frequency components via the wavelet transform (WT). Then, the infrared and visible features are exchanged through the Swap Channel operation, followed by an Average Pooling operation to obtain region-wise representations:
\begin{equation}
\begin{aligned}
&\mathbf{X}_{L}^{ir},\mathbf{X}_{H}^{ir}=WT(\mathbf{X}_{ir}^{*}),\mathbf{X}_{L}^{rgb},\mathbf{X}_{H}^{rgb}=WT(\mathbf{X}_{rgb}^{*})\\
&\mathbf{X}_{1},\mathbf{X}_{2} = avgpool(SC(\mathbf{X}_{L}^{ir},\mathbf{X}_{L}^{rgb}))
\end{aligned},
\end{equation}
Where $avgpool(\cdot)$ denotes the average pooling operation, and $SC(\cdot)$ denotes the swap channel operation using channels from the other modality in a ratio of $\alpha$. 
These representations are fed into the CMB module to extract cross-region saliency patterns:
\begin{equation}
\mathbf{w}_{ir},\mathbf{w}_{rgb} = \sigma(Expand(CMB(\mathbf{X}_{1},\mathbf{X}_{2}))),
\end{equation}
Where $\sigma(\cdot)$ denotes the sigmoid function, $Expand(\cdot)$ is spatial expansion to reshape, and $CMB(\cdot)$ is the application of the CMB module.
After being expanded and activated by sigmoid ($\sigma$), the generated weights act as gating factors, which are multiplied element-wise with the frequency components of each modality, thereby adaptively filtering modality-specific information. This frequency-domain gating enables the model to emphasize structurally consistent components and down-weight unreliable misaligned responses, without requiring explicit spatial alignment. Finally, the fused representation is reconstructed by the inverse wavelet transform (IWT) and refined by a $1 \times 1$ convolution to produce the final fused feature $\mathbf{X}^i_f$:
\begin{equation}
\begin{aligned}
&f=\mathbf{w}_{rgb} \otimes WT(\mathbf{X}_{rgb}^{*}) + \mathbf{w}_{ir} \otimes WT(X_{ir}^{*})\\
&\mathbf{X}_{f}^{i} = Conv_{1\times1}(IWT(f)).
\end{aligned}
\end{equation}
where $\otimes$ denotes element-wise multiplication and $Conv_{1\times1}(\cdot)$ represents a $1 \times 1$ convolution.
All three fused outputs \(\{\mathbf{X}_{f}^{1}, \mathbf{X}_{f}^{2}, \mathbf{X}_{f}^{3}\}\)are used as the multi-scale inputs to the YOLO detection head for final target prediction. In contrast to align-then-fuse approaches, this frequency-guided strategy achieves stable and complementary fusion by operating on shift-tolerant representations rather than fragile spatial correspondences. By leveraging region- and channel-wise joint attention, FCF effectively alleviates misalignment-induced feature interference while preserving discriminative and complementary information, thus improving the robustness and accuracy of misaligned multimodal feature fusion. This module is placed after VCC for feature fusion, as shown in Fig.~\ref{method}.

\begin{figure*}[htb]
    \centering
    \hspace{-1.0cm}
    \includegraphics[trim={100 0 275 0}, clip, width=\textwidth]{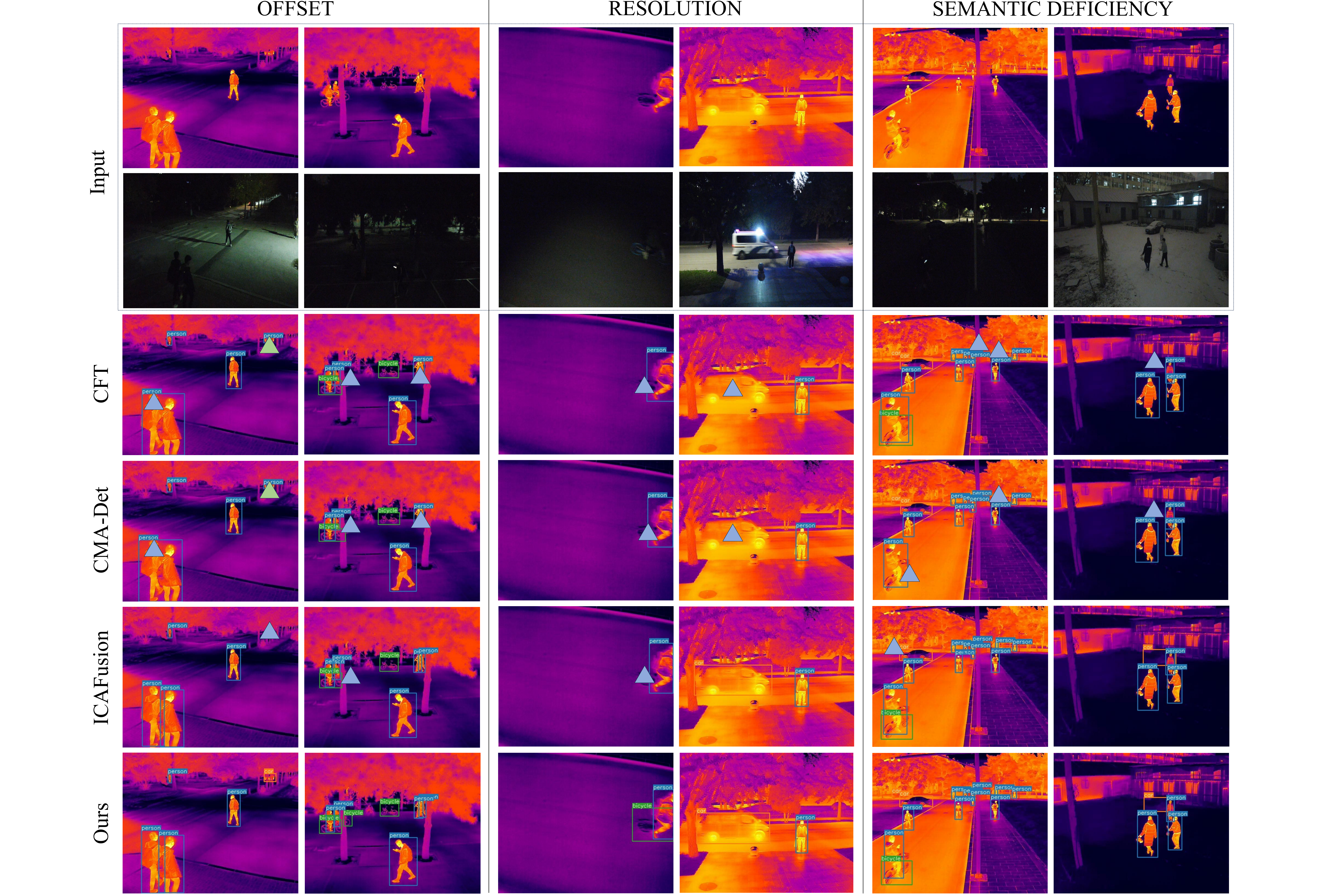}
    \vspace{-3 mm}
    \caption{Visualized comparison of our method to CFT, CMA-Det, and ICAFusion on the DVTOD dataset. \textcolor{blue}{BLUE} triangle labels missing targets. \textcolor{green}{GREEN} triangle labels false detection.}
    \label{compare}
\end{figure*}

\begin{table}[htbp]
\caption{Dataset Statics}
\label{dataset}
\centering
\setlength{\tabcolsep}{3pt}
\begin{tabular}{l|ccccc}
\toprule
\textbf{Dataset} & \textbf{Split} & \textbf{Images} & \textbf{Misaligned} & \textbf{Scene} & \textbf{Resolution}  \\ 
\midrule \midrule
\multirow{2}{*}{DVTOD}   & Train & 3212   & \multirow{2}{*}{Yes}        & \multirow{2}{*}{Drone} & RGB: $1920\times 1080$ \\
        &  val  & 1146   &            &       & IR:  $640\times 512$   \\ \midrule
\multirow{2}{*}{DroneVehicle}&Train& 35960& \multirow{2}{*}{Partly}     & \multirow{2}{*}{Drone} & \multirow{2}{*}{$640\times 512$}\\
        &  val  & 2938   &            &       &     \\ \midrule
\multirow{2}{*}{M$^3$FD} & Train & 5880   & \multirow{2}{*}{No}         & \multirow{2}{*}{Various} & \multirow{2}{*}{$1024\times 768$}\\
        &  val  & 2520   &            &       &     \\
\bottomrule
\end{tabular}
\end{table}

\section{Experiments}

\subsection{Experimental Settings}

\subsubsection{Datasets \& Metrics}
We conduct experiments on three challenging datasets: DVTOD~\cite{song2024misaligned}, and $M^3FD$~\cite{liu2022target}, DroneVehicle~\cite{sun2022drone} to evaluate our model.
They respectively represent three typical scenarios: misalignment with a large offset, resolution discrepancy, and semantic deficiency; spatially aligned data with semantic deficiency; and light offset data.
The DVTOD dataset comprises low-altitude drone scenarios, the $M^3FD$ dataset is captured from a vehicle-mounted perspective across various environments, and DroneVehicle contains large-scale high-altitude drone scenes.
Detailed statistics are provided in Table \ref{dataset}.
For quantitative evaluation, we use mean Average Precision metrics, including $mAP_{0.5}$ (at IoU=0.5) and $mAP$ (averaged over IoU thresholds from 0.5 to 0.95 with a step size of 0.05).

\subsubsection{Implementation Details}
The backbone network is built upon CFT~\cite{qingyun2021cross}, which serves as the baseline for our model.
The detector is optimized using the Stochastic Gradient Descent (SGD) algorithm. The number of training epochs is set to 300 for the DVTOD and $M^3FD$ datasets, and 150 for the DroneVehicle dataset.
A batch size of 4 is used, and all input images are resized to $640 \times 640$ pixels during training. We used pre-trained weights on COCO for the DVTOD dataset following \cite{song2024misaligned}.
All experiments were conducted on an NVIDIA GeForce RTX 4080S GPU.

\subsection{Comparison with State-of-the-Art Methods}

In this section, we compare our method with representative state-of-the-art methods on three benchmark datasets.
All methods are evaluated under consistent settings to ensure a fair and comprehensive assessment.

\begin{table*}[htbp]
\centering
\caption{Performance comparison of different methods on the DVTOD dataset.\\
The best result in each column is highlighted in \textbf{bold}, and the second-best is \underline{underlined}.}
\label{dvtodresults}
\normalsize
\begin{tabular}{c|ll||ccc|cccc}
\toprule
\multirow{2}{*}{\textbf{Modality}} & \multicolumn{1}{c}{\multirow{2}{*}{\textbf{Model}}} & \multirow{2}{*}{\textbf{Detector}} &  \textbf{Person} & \textbf{Car}  & \textbf{Bicycle} & $mAP_{0.5}$ & $mAP$ & \textbf{Param.} & \textbf{Speed}\\
 &  &  &  (\%)~$\uparrow$ & (\%)~$\uparrow$  & (\%)~$\uparrow$ & (\%)~$\uparrow$ & (\%)~$\uparrow$ & (M)~$\downarrow$ & (FPS)~$\uparrow$ \\
\midrule
\midrule
\multirow{3}{*}{IR Only} 
& YOLOv5-l  & YOLOv5  & 84.8 & 73.8 & 87.0 & 81.8 & 49.8 & - & -  \\
& YOLOv8-l  & YOLOv8  & 85.5 & 72.5 & 87.3 & 81.8 & 50.1 & - & -  \\
& YOLOv11-l & YOLOv11 & 85.1 & 72.1 & 86.5 & 81.2 & 49.4 & - & -  \\
\midrule
\multirow{9}{*}{IR+RGB} 
& CFT~\cite{qingyun2021cross}       & YOLOv5    & 89.0     & 78.1 & 81.4    & 82.8   & 47.7   & 207.86  & 75.58  \\
& CMX~\cite{zhang2023cmx}  & YOLOv5  & 88.9 & 75.9 & 79.6    & 81.6   & -   & 132.00 & - \\
& SuperYOLO~\cite{zhang2023superyolo} & YOLOv5    & 78.8   & 56.2 & 76.8    & 70.6   & 38.3 & \textbf{7.73}  & - \\
& CMA-Det~\cite{song2024misaligned}   & YOLOv5    & 90.4   & 81.6 & 83.0      & 85.0     & 46.7 & 33.33 & 83.50   \\
& ICAFusion~\cite{shen2024icafusion} & YOLOv5    & 90.3   & 81.2 & 83.4    & 85.0   & 48.7 & 120.21  & 92.50 \\
& GCMF-Net~\cite{qu2025gcmf} & YOLOv11    & \textbf{91.9}   & 81.2 & 83.7    & 85.6   & - & 54.31   & 51.76\\
& GAANet~\cite{zheng2025gaanet} & YOLOv5    & -   & - & -    & \textbf{88.0}   & \textbf{51.2} & 79.00 & 19.60 \\
\cmidrule(lr){2-10} 
& AlignFreeNet (\textbf{Ours})      & YOLOv5    & 91.2     & \textbf{85.5} & {\ul 86.6}    & 87.7   & {\ul 51.0} & 24.13 & {\ul 98.52}  \\
& AlignFreeNet (\textbf{Ours})      & YOLOv8    & {\ul 91.5}     & {\ul 85.0} & \textbf{87.5}    & \textbf{88.0}   & 50.2 & {\ul 23.74} & \textbf{101.21}  \\ \bottomrule
\end{tabular}
\end{table*}

\begin{table*}[htbp]
\caption{Performance Comparison of different methods on the $M^3FD$ dataset}
\label{m3fdresults}
\centering
\normalsize
\begin{tabular}{c|l||cc|cccccc} 
\toprule
\textbf{Modality} & \multicolumn{1}{c||}{\textbf{Model}}       & $mAP_{0.5}$     &  $mAP$ & \textbf{People} & \textbf{Bus} & \textbf{Car} & \textbf{Motorcycle} & \textbf{Lamp} & \textbf{Truck}  \\ \midrule \midrule
RGB Only&YOLOv8 & 80.9 & 52.5 & 70.6 & 92.9 & 91.2 & 69.9 & 75.3 & 86.0 \\
IR Only&YOLOv8 & 79.5 & 53.1 & 82.9 & 90.9 & 90.0 & 64.6 & 63.0 & 85.9 \\ \midrule
\multirow{10}{*}{RGB+IR}
&TarDAL~\cite{liu2022target} & 80.5 & 54.1 & 81.5 & 81.3 & \textbf{94.8} & 69.3 & 87.1 & 68.7 \\
&SuperFusion~\cite{tang2022superfusion} & 83.5 & 56.0 & {\ul 83.7} & 93.2 & 91.0 & 77.4 & 70.0 & 85.8  \\
&CDDFuse~\cite{zhao2023cddfuse} & 81.1 & 54.3 & 81.6 & 82.6 & 92.5 & 71.6 & 86.9 & 71.5 \\
&IGNet~\cite{li2023learning} & 81.5 & 54.5 & 81.6 & 82.4 & 92.8 & 73.0 & 86.9 & 72.1 \\
&CMADet~\cite{song2024misaligned} & 80.6 & 49.2 & 80.2 & 89.3 & 90.5 & 68.7 & 71.8 & 82.4 \\
&FusionMamba~\cite{dong2024fusion} & 85.0 & 57.5 & 80.3 & 92.8 & 91.9 & 73.0 & 84.8 & 87.1 \\
&MMFN~\cite{yang2024multidimensional} &86.2&57.4 & 83.0 & 92.1 & 93.2 & 73.7 & 87.6 & 87.4 \\
&EMMA~\cite{zhao2024equivariant} & 82.9 & 55.4 & 82.0 & 83.2 & 93.5 & {\ul 77.7} & {\ul 87.7} & 73.5  \\
&RetinexDet-B~\cite{liu2025retinexdet} & {\ul 86.4} &\textbf{61.9} & \textbf{87.5} & {\ul 93.3} & {\ul 93.8} & 76.2 & 80.8 & {\ul 86.8} \\
\cmidrule(lr){2-10}
&AlignFreeNet (\textbf{Ours}) & \textbf{88.3} & {\ul 57.5} & 83.6 & \textbf{95.1} & 93.2 & \textbf{81.0} & \textbf{88.3} & \textbf{88.6} \\
\bottomrule
\end{tabular}
\end{table*}

\subsubsection{DVTOD}

As shown in Table \ref{dvtodresults}, we compare the performance of different models on the DVTOD dataset.
Due to modality misalignment, simply increasing the backbone size does not guarantee better performance in cross-modality detection.
Our experiments show that the fusion strategy plays a crucial role.
SuperYOLO~\cite{zhang2023superyolo}, for instance, performs early-stage fusion, which propagates misaligned features and significantly degrades performance.
CMA-Det~\cite{song2024misaligned} performs competitively on DVTOD, but it struggles to generalize across datasets and shows no improvement when using larger backbones, resulting only in increased model size.
In contrast, our model achieves superior performance, reaching 88.0 in $\mathrm{mAP}_{0.5}$ and 50.2 in $\mathrm{mAP}$, and compared to the previous SOTA method, we reduced the model size by 70\% and run over $5\times$ faster.
Qualitative visualizations are presented in Fig. \ref{compare}.
The original single-modal YOLOs commonly have a higher precision in the Bicycle genre than the multi-modal methods.
The cross-modality alignment-fusion method usually blurs the boundary of targets, especially small targets like bicycles.
In a low-light environment, RGB images often fail to clearly depict them, and infrared imaging also struggles due to their low thermal signature, like the one illustrated in Fig. \ref{compare}, while single-modal structure can still extract them, fusion methods frequently cause them to blend into the background.
For the People category, misalignment and deficiencies in existing alignment strategies often cause adjacent targets to merge. As shown in the OFFSET part of the figure, other methods frequently detect close-standing individuals as a single target. In contrast, our wavelet structure preserves clear edges and boundaries, yielding a 91.5 $mAP_{0.5}$ for this category. Additionally, our variation-based selective fusion strategy effectively suppresses redundant cross-modal features, reducing representation ambiguity.
For the most misaligned category, Car, our method achieves an $mAP_{0.5}$ of 85.0, demonstrating its robustness to severe spatial misalignment.

\subsubsection{$M^3FD$}
To further assess the self-adaptiveness of our method, we conducted experiments on the $M^3FD$ dataset, which comprises a diverse range of scenes. Due to the detection set not being officially split, we follow the split of the training and validation sets in \cite{dong2024fusion}. 
We compared our method against nine cross-modality methods: TarDal, SuperFusion, CDDFuse, IGNet, CMA-Det, FusionMamba, MMFN, EMMA, and RetinexDet-B.
As shown in Table \ref{m3fdresults}, our method achieves $88.3\%$ in the $mAP_{0.5}$ metric and $57.5\%$ in overall $mAP$, outperforming all other methods.
These results validate the self-adaptive nature of our method, which remains robust in non-offset scenarios that contain only semantic deficiencies.

\begin{figure*}[t!]
    \centering
    \hspace{0cm}
    \includegraphics[trim={75 450 75 350}, clip, width=\textwidth]{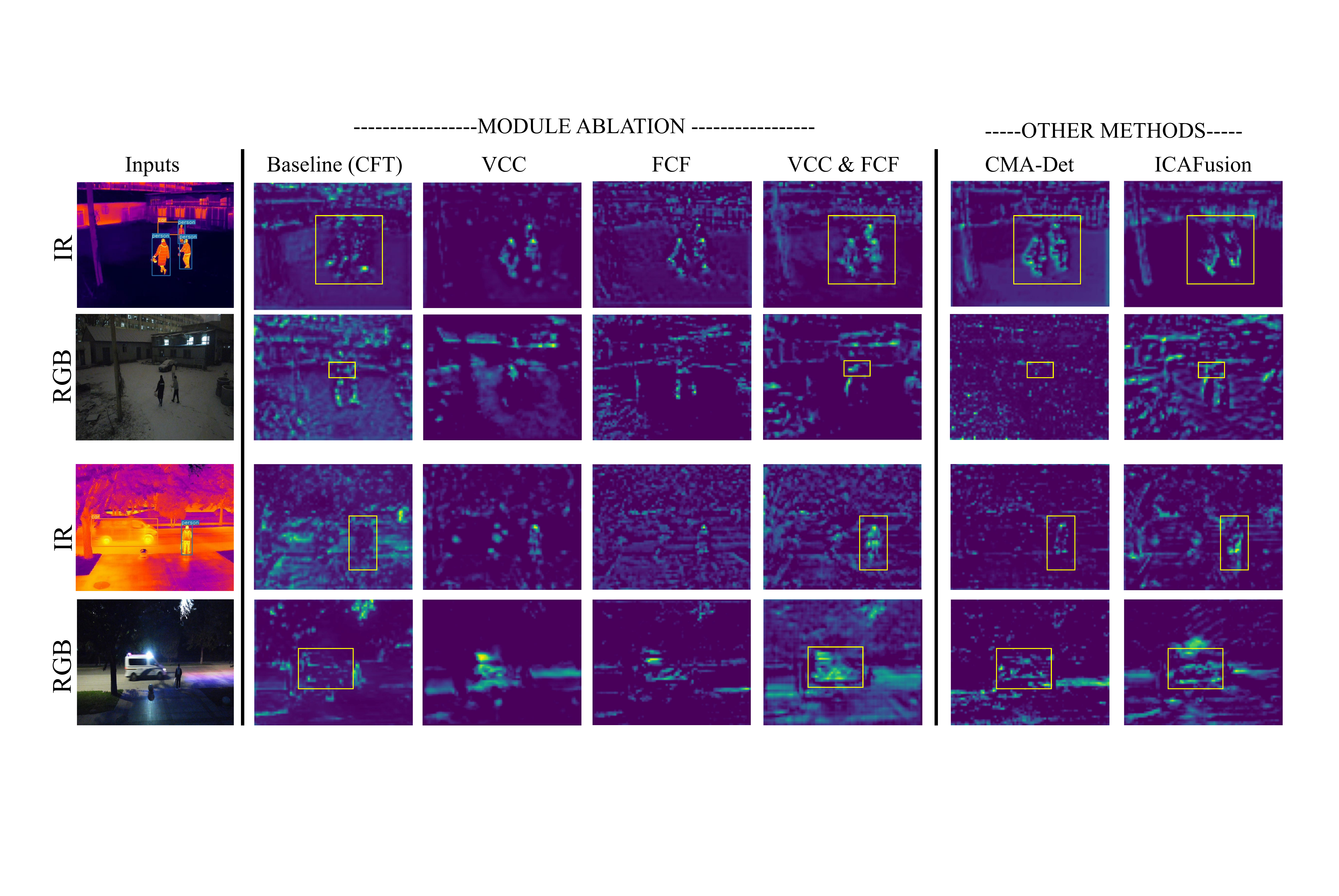}
    \vspace{-8 mm}
    \caption{Visualized comparison of feature maps for ablation study on the DVTOD dataset. \textcolor{yellow}{YELLOW} box labels target features.}
    \label{ablationcomparison}
\end{figure*}

\begin{table}[htbp]
\caption{Performance comparison of different methods on the DroneVehicle dataset (HBB)}
\label{droneresults}
\centering
\normalsize
\setlength{\tabcolsep}{6pt}
\begin{tabular}{l|c|c|c}
\toprule
\textbf{Method} & \textbf{Modality} & $\mathbf{mAP_{0.5}}$ & $\mathbf{mAP}$ \\
\midrule
YOLOv5-s & RGB Only & 74.6 & 46.7 \\
YOLOv5-s & IR Only & 80.8 & 60.2 \\
\midrule
CFT~\cite{qingyun2021cross} & \multirow{7}{*}{RGB+IR} & 84.3 & 61.2 \\
SuperYOLO~\cite{zhang2023superyolo} &  & 81.4 & 58.6 \\
GHOST~\cite{zhang2023guided} &  & 81.5 & 59.3 \\
ICA-Fusion~\cite{shen2024icafusion} &  & {\ul 84.8} & {\ul 62.1} \\
GM-DETR~\cite{xiao2024gm} &  & 80.8 & 55.9 \\
CMA-Det~\cite{song2024misaligned} &  & 82.0 & 59.5 \\
\textbf{AlignFreeNet (Ours)} &  & \textbf{85.8} & \textbf{63.1} \\
\bottomrule
\end{tabular}
\end{table}


\subsubsection{DroneVehicle}
We evaluate the proposed method against six state-of-the-art methods on the DroneVehicle dataset.
As shown in Table \ref{droneresults}, our method achieves the highest performance in terms of both $mAP_{0.5}$ and overall $mAP$.
Owing to the inherent spatial offset between the RGB and infrared modalities, the annotations for the two modalities are not perfectly aligned.
For consistency, we adopt the infrared annotations as the ground truth during both training and evaluation.
Method introduced by Song et al.~\cite{song2024misaligned}, was specifically designed to address severe cross-modal misalignment, such as that found in the DVTOD dataset.
However, when applied to scenarios with relatively minor misalignment, its performance deteriorates.
In such cases, its alignment mechanism may in fact impede accurate detection.
By contrast, our method demonstrates strong adaptability to varying levels of misalignment and achieves superior results across all benchmarks.

\begin{figure*}[htbp]
    \centering
    \hspace{0cm}
    \includegraphics[trim={0 200 0 250}, clip, width=\textwidth]{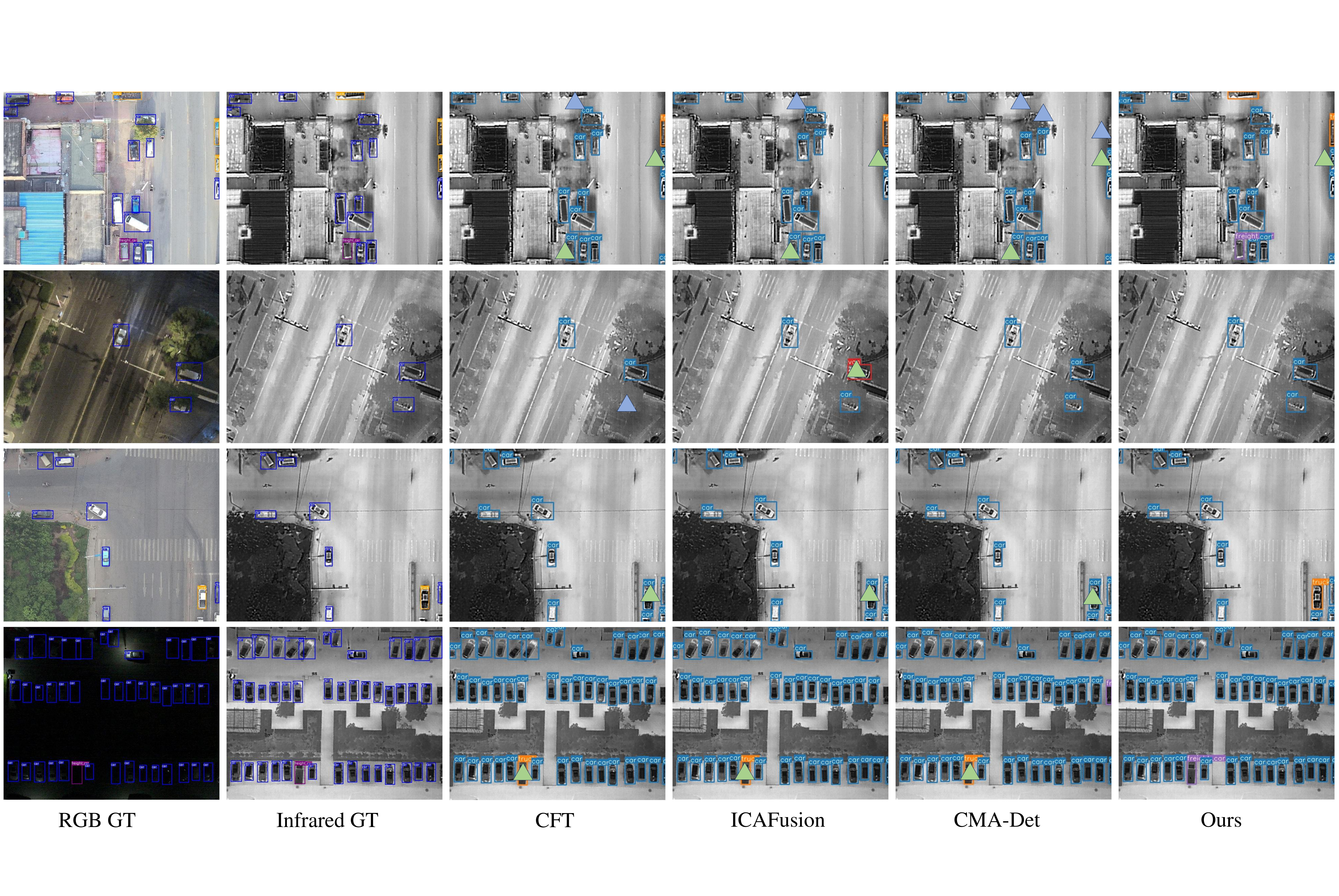}
    \caption{Visualized comparison of our method to CFT, CMA-Det, ICAFusion on the DroneVehicle dataset. \textcolor{blue}{BLUE} triangle labels missing targets. \textcolor{green}{GREEN} triangle labels false detection.}
    \label{compare2}
\end{figure*}

\subsection{VCC Number and Position Analysis}
Since there are five possible positions for placing the VCC module within the network, as shown in Fig.\ref{method}, the impact of varying both the number and placement of VCCs remains to be explored. Therefore, in this section, we design a series of experiments to determine the optimal configuration. As demonstrated in Table \ref{VCCAnalysis}, the VCC module is more suitable for processing low- and mid-level features, achieving its best performance when placed at \{P2, P3\}.

\begin{table}[htbp]
    \caption{Effect of the Number and Position of VCC Modules}
    \label{VCCAnalysis}
    \normalsize
    \setlength{\tabcolsep}{4pt}
    \centering
    \begin{tabular}{c||ccc}
    \toprule
    \textbf{Position and Number} & $mAP_{0.5}$ & $mAP$ & \textbf{Param.}\\
    \midrule
    P2   & 87.1   & 49.2 & \textbf{22.29}     \\
    P2,P3   & \textbf{88.0}   & 50.2 & 23.74     \\
    P3,P4   & 87.2   & 50.1 & 28.63     \\
    P2,P3,P4   & 87.8   & \textbf{50.6} & 29.06     \\
    P2,P3,P4,P5   & 87.0   & 50.0 & 49.39     \\
    \bottomrule
    \end{tabular}
\end{table}

\subsection{Ablation Study}
In this section, we present ablation studies conducted on the DVTOD dataset. The results are summarized in Table~\ref{ablation}.
Our baseline is the CFT method.

\textbf{VCC}:
Despite consisting of only 1.84M parameters, VCC delivers notable improvements.
It mitigates performance degradation caused by misalignment and enhances complementary modality variation.
When compared with the baseline using Transformer, VCC improves $mAP_{0.5}$ by 4.1\%, the parameters are reduced by 92.5\%, and the speed almost doubled.

\textbf{FCF}:
This module is designed to complement late-stage fusion mechanisms.
Compared to the baseline, FCF improves $mAP_{0.5}$ by 4.2\%, the parameters are reduced by 89.5\%, and the speed is increased by 79.4\%.

A visualization of the feature maps for the ablation study is shown in  Fig.~\ref{ablationcomparison} to demonstrate the effectiveness of each module and to do a comparison with other models.

\begin{table}[htbp]
    \caption{Ablation study on the DVTOD dataset}
    \label{ablation}
    \normalsize
    \setlength{\tabcolsep}{4pt}
    \centering
    \begin{tabular}{c|cc||ccc}
    \toprule
    \multicolumn{3}{c||}{\textbf{Components}} & \multirow{2}{*}{$mAP_{0.5}$} & \multirow{2}{*}{\textbf{Param.}} & \multirow{2}{*}{\textbf{Speed}}\\
    \textit{Transformer} & \textit{VCC} & \textit{FCF} & & & \\
    \midrule
    \midrule
        \checkmark &    &    & 82.8   & 207.86 & 75.6     \\
         &  \checkmark  &    & 86.9   & \textbf{15.97} & \textbf{145.72}     \\
         &    &  \checkmark  & 87.0   & 21.86 & 135.59     \\
         &  \checkmark  &  \checkmark  & \textbf{88.0}   & 23.74 & 101.2    \\
    \bottomrule
    \end{tabular}
\end{table}

\section{Conclusion}

In this work, we present AlignFreeNet, a unified light-weight alignment-free network for VI-OD under diverse and challenging misalignment conditions.
Addressing a long-standing issue in multi-modal detection, AlignFreeNet offers a principled solution that leverages frequency-awareness and cross-modal variation adaptation to mitigate spatial misalignment without reliance on strictly aligned supervision.
Our design introduces core innovations: VCC and FCF.
These modules work synergistically to model features in the hybrid spatial-frequency domain, promoting reliable cross-modal fusion and suppressing cross-modal interference.
Comprehensive experiments across three benchmarks demonstrate that AlignFreeNet consistently outperforms existing methods across both aligned and misaligned scenarios, achieving a superior balance between performance and efficiency.
Beyond empirical performance, AlignFreeNet represents a generalizable insight that opens new possibilities for alignment-free fusion perception of VI-OD.
Looking ahead, we plan to extend AlignFreeNet to additional modalities and integrate it with lightweight deployment strategies for diverse degradation scenarios, thereby reducing the dependence on strictly aligned annotations in cross-modal perception tasks.

\bibliographystyle{IEEEtran}
\bibliography{acmart}

\begin{IEEEbiography}[{\includegraphics[width=1in,height=1.25in,clip,keepaspectratio]{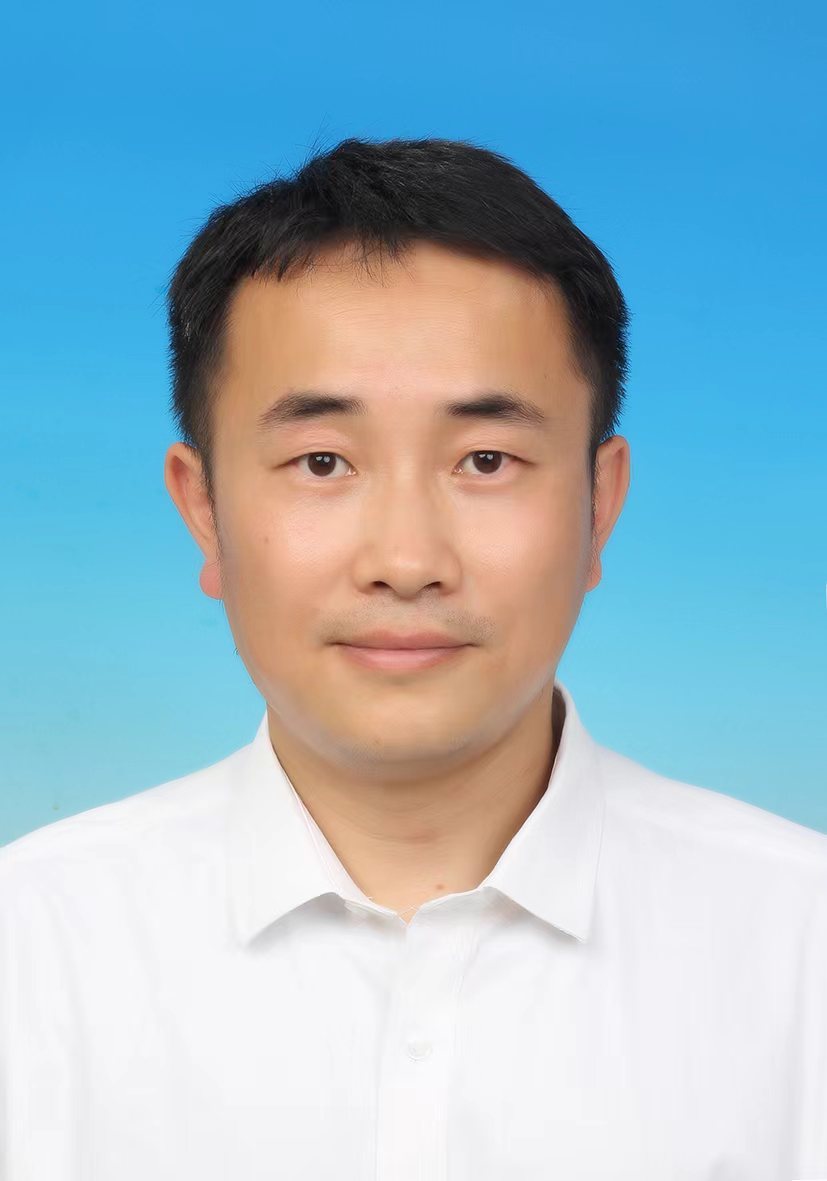}}]{Dingkun Zhu}
received the Ph.D. degree in the Department of Computer Science and Engineering,
Hong Kong Metropolitan University in 2023. He is currently an Associate Professor with the School of Computer Engineering, Jiangsu University of Technology. Before joining JSUT, he served as a postdoctoral fellow at The Hong Kong Polytechnic University. His research interests focus on geometry processing, computer graphics, and deep learning.
\end{IEEEbiography}

\vspace{-15 mm} 

\begin{IEEEbiography}[{\includegraphics[width=1in,height=1.25in,clip,keepaspectratio]{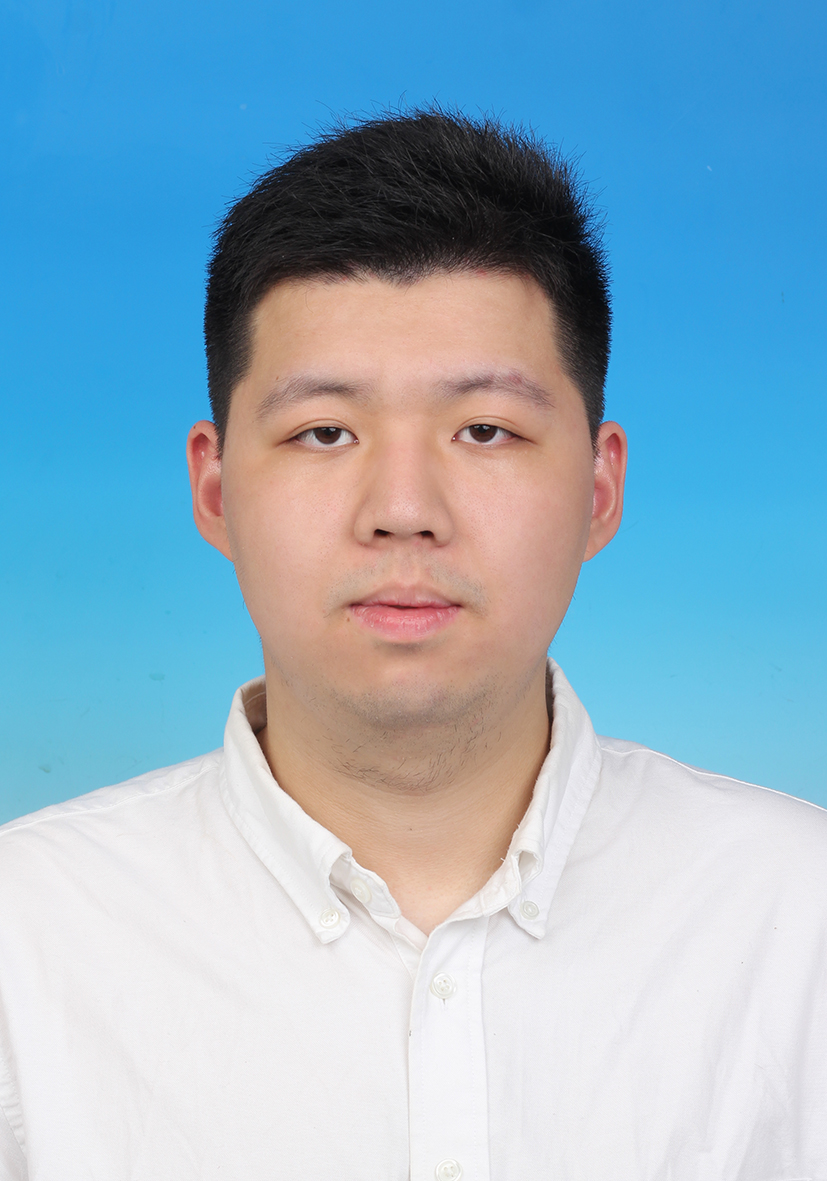}}]{Haote Zhang}
   is currently pursuing the PhD degree in the school of Computer Engineering, Jiangsu University of Technology. His research interests include computer vision, pattern recognition, and remote sensing.
\end{IEEEbiography}

\vspace{-15 mm} 

\begin{IEEEbiography}[{\includegraphics[width=1in,height=1.25in,clip,keepaspectratio]{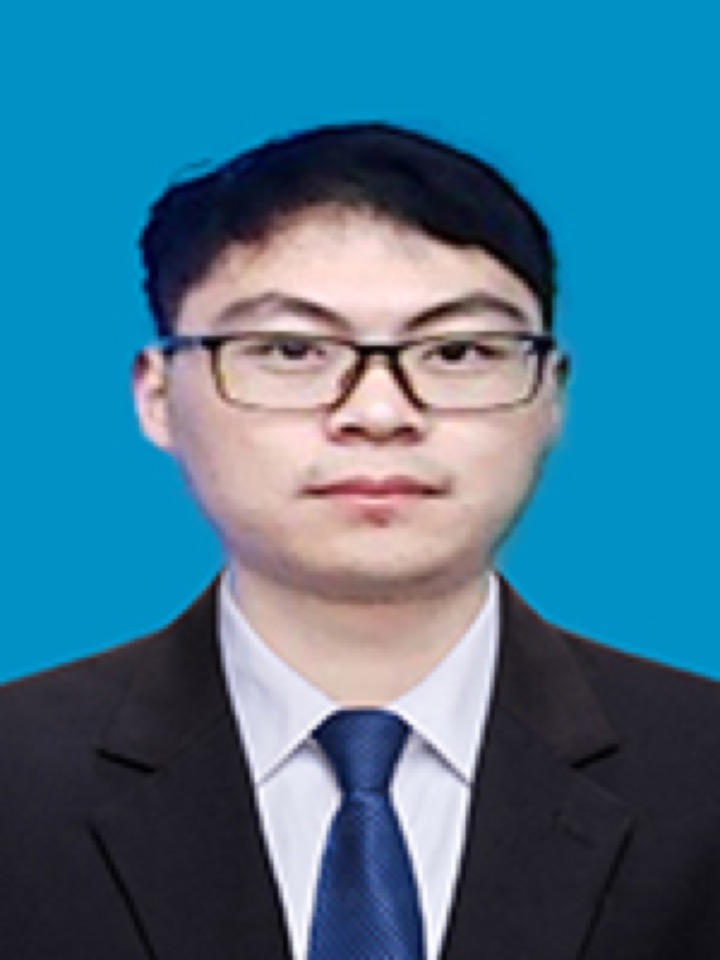}}]{Lipeng Gu}
is currently pursuing his Ph.D. degree with computer science and technology at Nanjing University of Aeronautics and Astronautics. He received the M.Sc. degree from Donghua University in 2021. His research interests include deep learning, computer vision, and computer graphics.
\end{IEEEbiography}

\vspace{-15 mm} 

\begin{IEEEbiography}[{\includegraphics[width=1in,height=1.25in,clip,keepaspectratio]{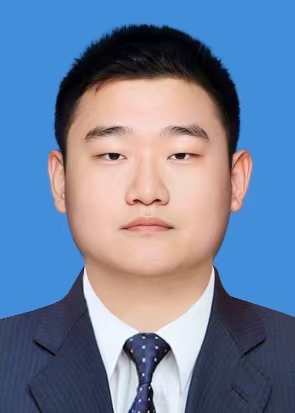}}]{Wuzhou Quan} received the B.E. degree in Electronic Information Engineering from Nanjing University of Posts and Telecommunications, China, in 2018, and the M.E. degree in Electronic Information from Shandong Technology and Business University, China, in 2024.
He is currently pursuing the Ph.D. degree in Computer Science and Technology at Nanjing University of Aeronautics and Astronautics, China.  
His research interests include computer vision, pattern recognition, and remote sensing.
\end{IEEEbiography}

\vspace{-15 mm} 

\begin{IEEEbiography}[{\includegraphics[width=1in,height=1.25in, clip,keepaspectratio]{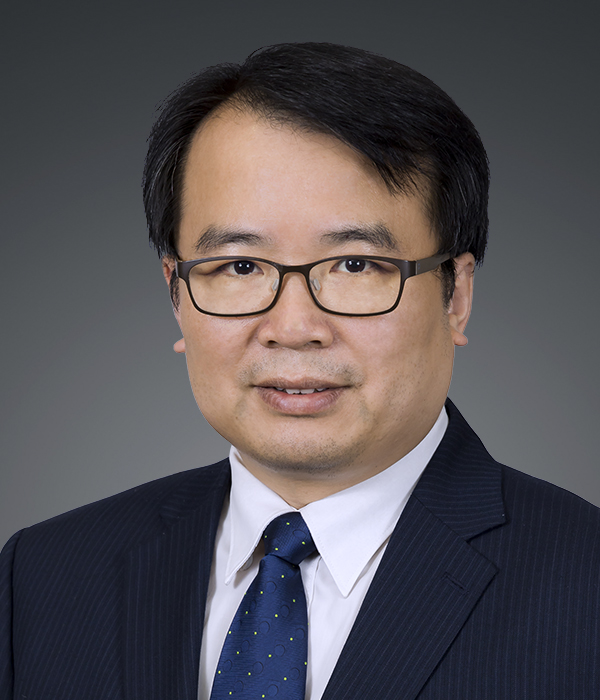}}]{Fu Lee Wang} (Senior Member, IEEE) received the B.Eng. degree in computer engineering and the MPhil. degree in computer science and information systems from The University of Hong Kong, and the Ph.D. degree in systems engineering and engineering management from the Chinese University of Hong Kong. Prof. Wang is the Dean of the School of Science and Technology, Hong Kong Metropolitan University. He is a fellow of BCS, HKIE and IET and a Senior Member of ACM. He was the Chair of the IEEE Hong Kong Section Computer Chapter and ACM Hong Kong Chapter.
\end{IEEEbiography}

\vspace{-15 mm} 

\begin{IEEEbiography}[{\includegraphics[width=1in,height=1.25in,clip,keepaspectratio]{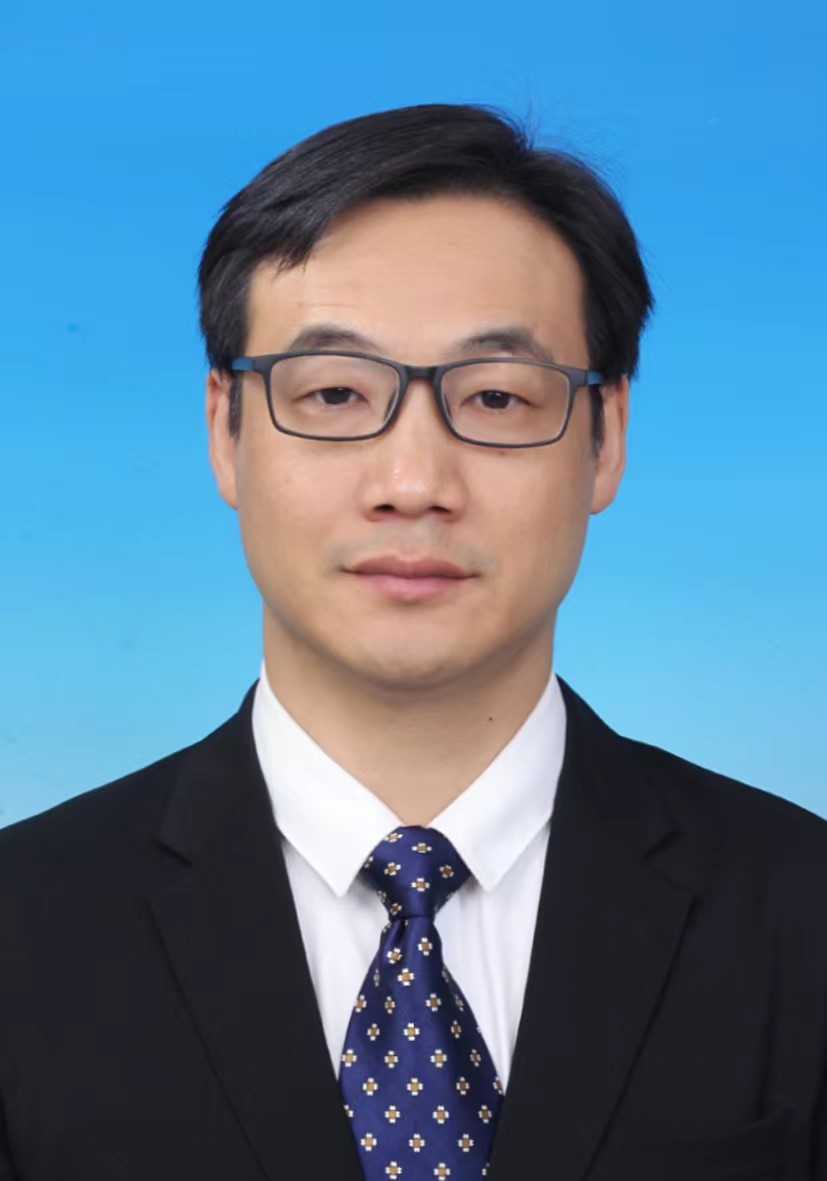}}]{Honghui Fan} received the Ph.D. degree in Yamgata University of Japan, in 2011. He is currently a Professor with the School of Computer Engineering, Jiangsu University of Technology. His current research interests include image processing, computer vision, image reconstruction, and image restoration.
\end{IEEEbiography}

\vspace{-15 mm} 

\begin{IEEEbiography}[{\includegraphics[width=1in,height=1.25in,clip,keepaspectratio]{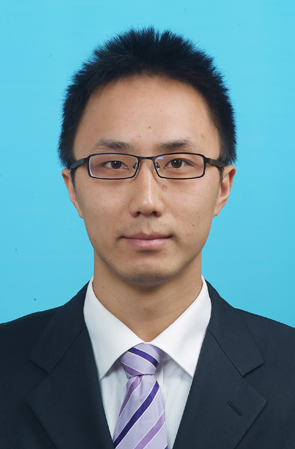}}]{Jiali Tang} received the Ph.D. degree in mechatronic engineering from Jiangsu University, China,
in 2016. He is currently a Professor with the
School of Computer Engineering, Jiangsu University
of Technology. His research interests include
image processing, computer vision, and pattern
recognition.
\end{IEEEbiography}

\vspace{-15 mm}

\begin{IEEEbiography}
[{\includegraphics[width=1in,height=1.25in, clip,keepaspectratio]{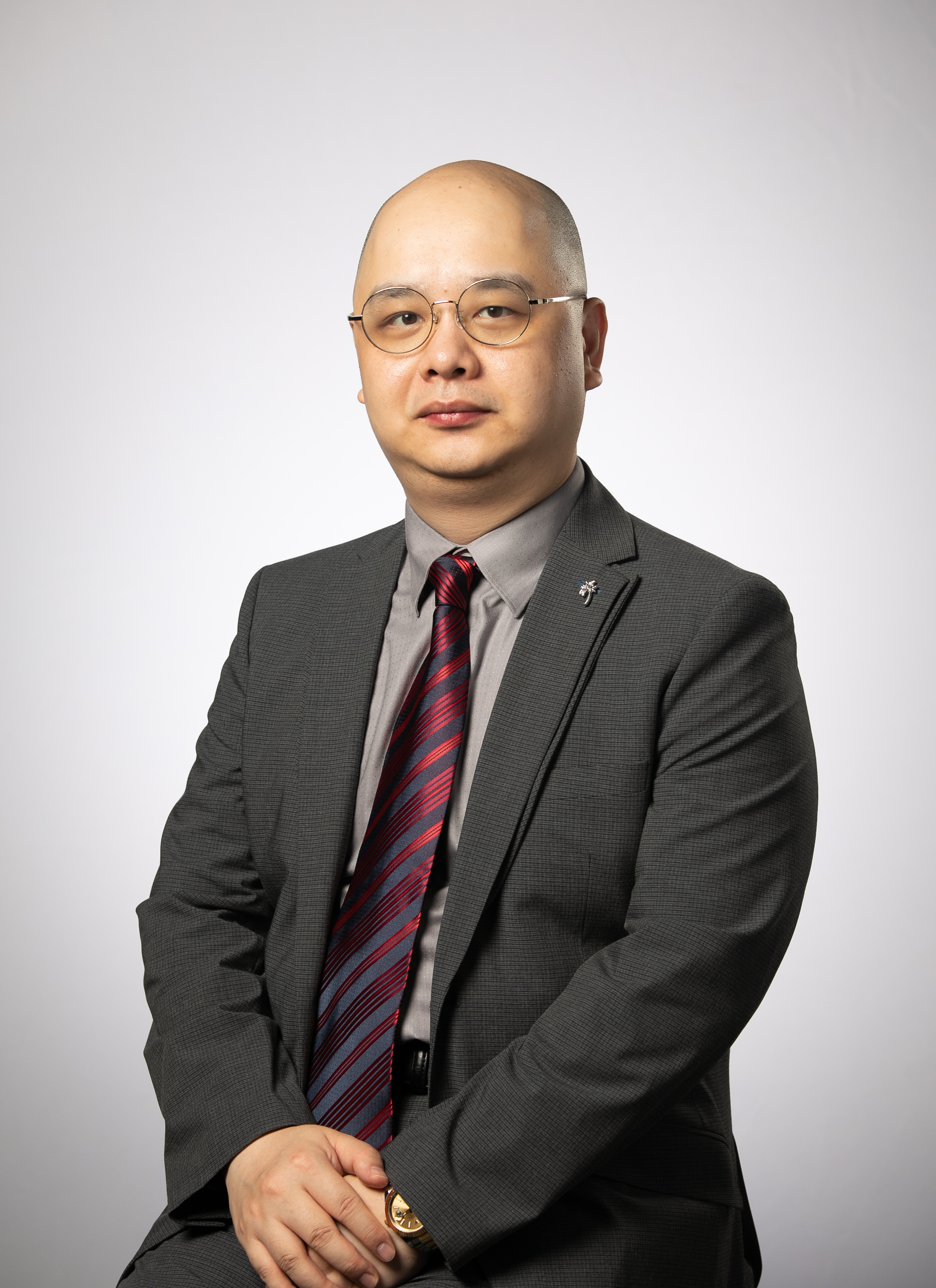}}]{Haoran Xie} (Senior Member, IEEE) received a Ph.D. degree in Computer Science from City University of Hong Kong and an Ed.D degree in Digital Learning from the University of Bristol. He is currently the Department Head and Associate Professor at the Department of Computing and Decision Sciences, Lingnan University, Hong Kong. He serves as the Editor-in-Chief of the Natural Language Processing Journal, Computers \& Education: Artificial Intelligence, and Computers \& Education: X Reality. He has been listed as one of the World's Top 2\% Scientists by Stanford University.
\end{IEEEbiography}

\vspace{-15 mm} 

\begin{IEEEbiography}[{\includegraphics[width=1in,height=1.25in, clip,keepaspectratio]{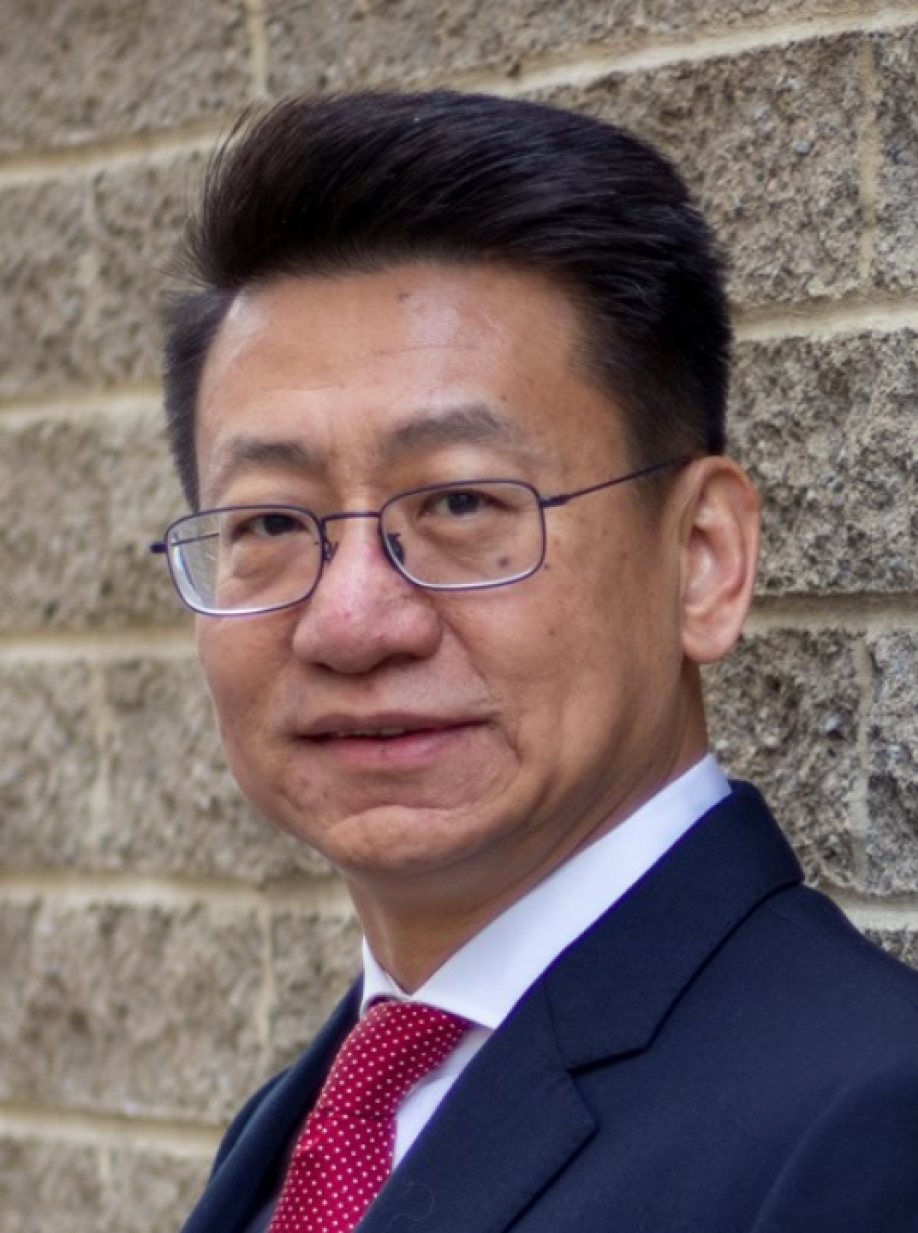}}]{Xiao-Ping Zhang} (Fellow, IEEE) received B.S. and Ph.D. degrees from Tsinghua University in 1992 and 1996, respectively, both in Electronic Engineering. He holds an MBA in Finance, Economics and Entrepreneurship with Honors from the University of Chicago Booth School of Business, Chicago, IL. 

He is a Fellow of the Canadian Academy of Engineering, Fellow of the Engineering Institute of Canada, Fellow of the IEEE, a registered Professional Engineer in Ontario, Canada, and a member of Beta Gamma Sigma Honor Society.  He is Editor-in-Chief for the IEEE Journal of Selected Topics in Signal Processing. He is the Senior Area Editor for the IEEE Transactions on Image Processing. He served as Senior Area Editor for the IEEE Transactions on Signal Processing and Associate Editor for the IEEE Transactions on Image Processing, the IEEE Transactions on Multimedia, the IEEE Transactions on Circuits and Systems for Video Technology, the IEEE Transactions on Signal Processing, and the IEEE Signal Processing Letters. He was selected as an IEEE Distinguished Lecturer by the IEEE Signal Processing Society and by the IEEE Circuits and Systems Society.  
\end{IEEEbiography}

\vspace{-15 mm} 

\begin{IEEEbiography}[{\includegraphics[width=1in,height=1.25in, clip,keepaspectratio]{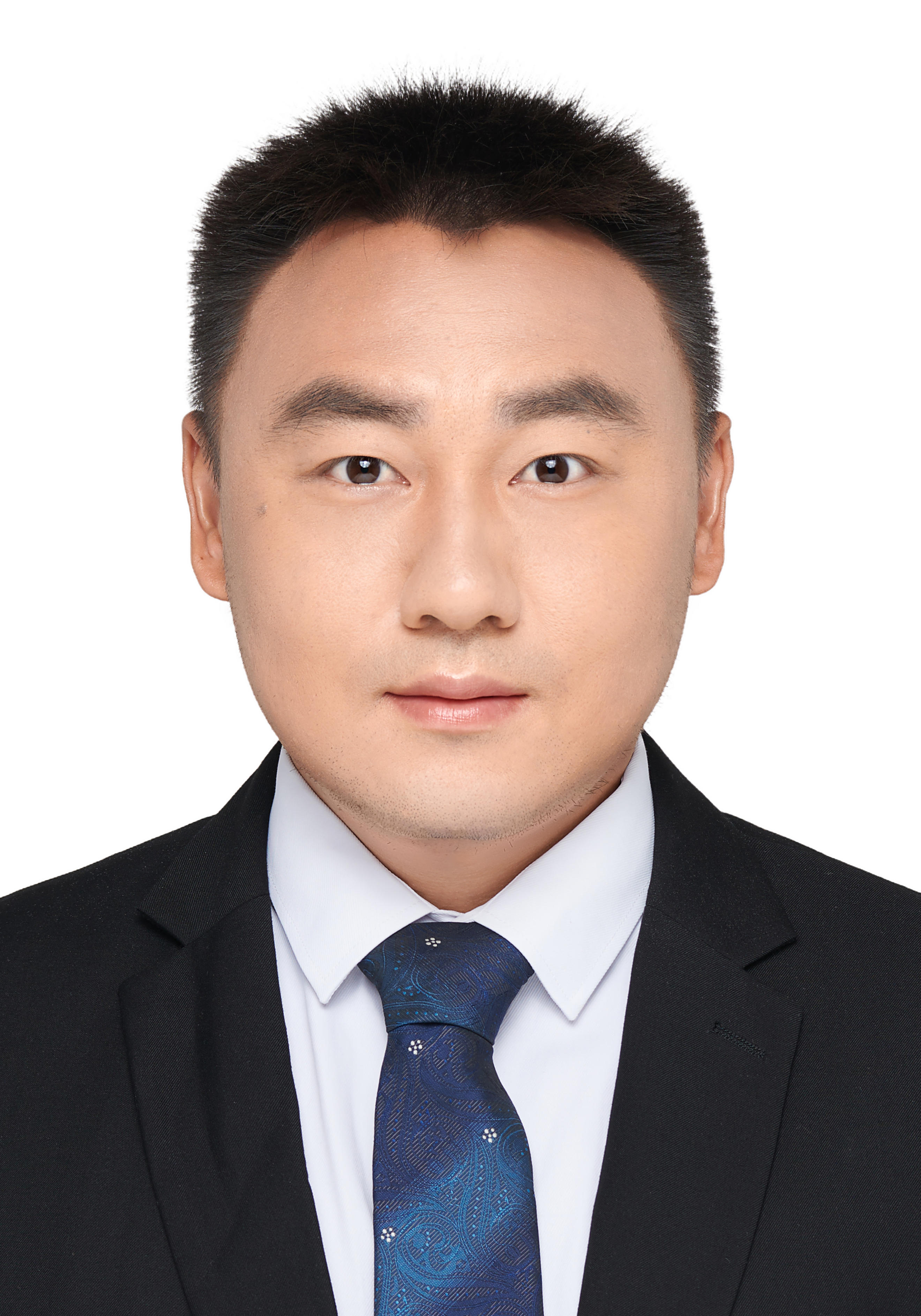}}]{Mingqiang Wei} (Senior Member, IEEE) received his Ph.D. degree (2014) in computer science and engineering from the Chinese University of Hong Kong (CUHK). He is a professor at the School of Computer Science and Technology, Nanjing University of Aeronautics and Astronautics (NUAA). Before joining NUAA, he served as an assistant professor at Hefei University of Technology and a postdoctoral fellow at CUHK. He is now an associate editor for IEEE Transactions on Image Processing, IEEE Transactions on Geoscience and Remote Sensing, ACM TOMM, and a Guest Editor for IEEE Transactions on Multimedia. His research interests focus on 3D vision, computer graphics, and deep learning.
\end{IEEEbiography}


\end{document}